\documentclass[letterpaper, 10 pt, conference]{ieeeconf}  

\IEEEoverridecommandlockouts                              

\usepackage{tikz}

\usepackage{times}
\usepackage{graphicx}
\DeclareGraphicsExtensions{.pdf,.png,.jpg}
\usepackage{amsmath}
\DeclareMathOperator*{\argmax}{argmax} 
\DeclareMathOperator*{\argmin}{argmin} 
\usepackage{amssymb}  
\usepackage{makecell}
\usepackage{booktabs}
\usepackage{multirow}
\usepackage{here}
\usepackage{kotex}

\title{\LARGE \bf
A Variational Feature Encoding Method of 3D Object for Probabilistic Semantic SLAM
}

\author{H. W. Yu and B. H. Lee
\thanks{*This work was supported by the National Research Foundation of Korea(NRF) grant funded by the Korea government(MSIP) (No. 2017R1A2B2002608), and in part by Automation and Systems Research Institute (ASRI)}
\thanks{H. W. Yu and B. H. Lee are with Automation and Systems Research Institute, Department of Electrical and Computer Engineering, 
	Seoul National University, Seoul, Korea(Republic of)
	{\tt\small \{bgus2000, bhlee\}@snu.ac.kr}}%
}

\begin{document}

\maketitle
\thispagestyle{empty}
\pagestyle{empty}

\begin{abstract}
This paper presents a feature encoding method of complex 3D objects for high-level semantic features.
Recent approaches to object recognition methods become important for semantic simultaneous localization and mapping (SLAM).
However, there is a lack of consideration of the probabilistic observation model for 3D objects, as the shape of a 3D object basically follows a complex probability distribution. 
Furthermore, since the mobile robot equipped with a range sensor observes only a single view, much information of the object shape is discarded. 
These limitations are the major obstacles to semantic SLAM and view-independent loop closure using 3D object shapes as features.
In order to enable the numerical analysis for the Bayesian inference, we  approximate the true observation model of 3D objects to tractable distributions.
Since the observation likelihood can be obtained from the generative model,
we formulate the true generative model for 3D object with the Bayesian networks.
To capture these complex distributions, we apply a variational auto-encoder.
To analyze the approximated distributions and encoded features, we perform classification with maximum likelihood estimation and shape retrieval.
\end{abstract}

\section{INTRODUCTION}
In robotics, simultaneous localization and mapping (SLAM) with high-level semantic features has become an important factor for scene understanding \cite{slam++}. 
The object oriented feature is ideal for view-independent loop-closure in SLAM, and is easily used for reinforcement learning such as object search \cite{object_search,RL_auxiliary}.
Especially, object recognition is widely used to capture the meaningful features and infer categories and viewpoints \cite{vggnet, rotationnet}.  
However, there is a lack of consideration of the probabilistic observation model for the object.
Recent studies for object recognition can be applied to object detection and evaluation of the classification probability \cite{faster_rcnn, yolo9000}.
Nevertheless, these methods concentrate on the detecting objects precisely for a single image, rather than estimating the probability distribution of the object. Therefore, it is cumbersome to apply it to the probabilistic SLAM which requires a tractable observation model capable of numerical analysis.
Intractable probabilistic distribution of the object’s shape also makes it difficult to estimate the observation model. 
Moreover, since 3D scene observed by mono camera or range finder is a single view, a mobile robot can observe only a part of an object at a glance. As an object has various single views, ideally, scanning the entire 3D shape of an object should be possible to achieve view-independent feature matching, loop-closure and complete volumetric maps. 
However, it is hard to obtain full shapes of objects while a mobile robot performing various tasks in real-time.

\begin{figure}[t]
	\centering
	\includegraphics[scale=0.075]{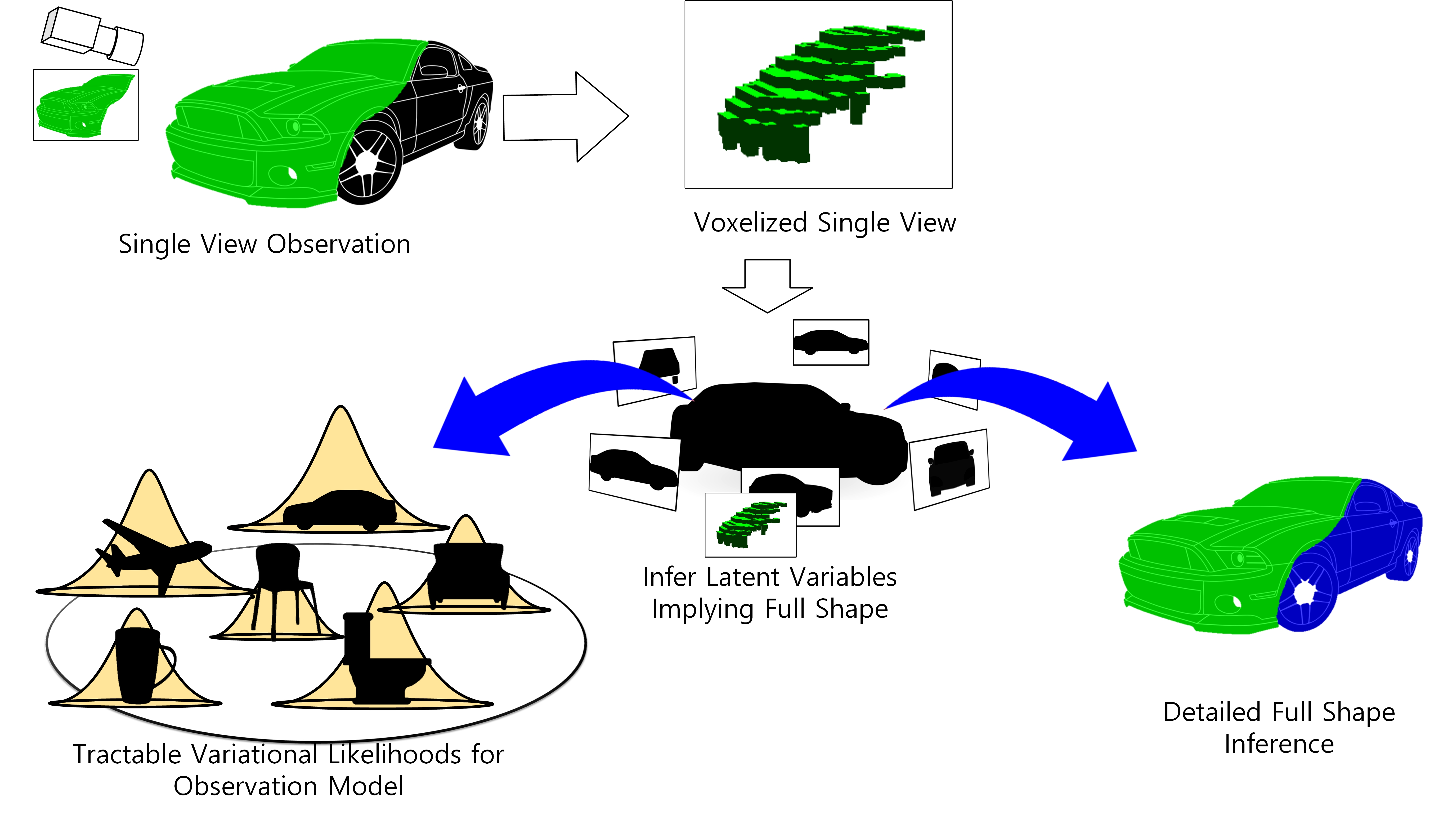}
	\caption{
		Overview of the proposed method. We represent an observed view data in voxel form. In order to approximate the intractable observation model of the 3D object, variational auto-encoder is used by learning the latent variables of the object shapes. Therefore, it is possible to perform numerical analysis of the semantic SLAM with 3D objects.
	}
	\label{likelihood_inference}
\end{figure}
\begin{figure*}[t]
	\centering
	\includegraphics[scale=0.20]{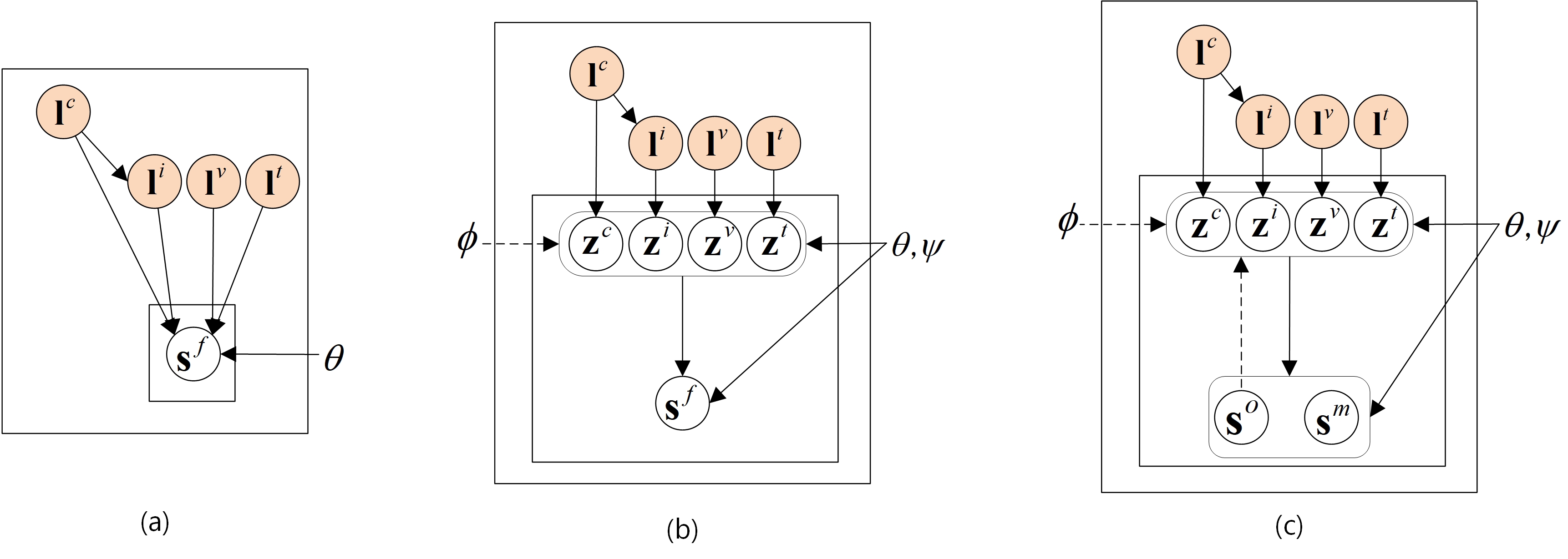}
	\caption{
		Proposed Bayesian process for variational approximation. (a) We first assume that the full shape of an object $\boldsymbol{s}^f$ is generated from the category $\boldsymbol{l}^c$ which roughly gives common  shape information for the class, and the instance label $\boldsymbol{l}^i$ which represents a detailed interpretation of the object. $\boldsymbol{l}^v$ and $\boldsymbol{l}^t$ are the labels of the observation viewpoint and the translation, respectively. (b) To capture the complex likelihood distribution, the variational latent variables $\boldsymbol{z}$ are adopted. 
		Priors for the latent variables are defined as conditional normal distributions, which also should be inferred from the data. (c) Since scanning the 3D full shape is challenging for a mobile robot, we utilize the observed single views to infer the variational latent variables. 
		Solid lines represent the generative model $p_{\theta,\psi}\left(\boldsymbol{s}^f,\boldsymbol{z} | \boldsymbol{l}\right) = p_\psi\left(\boldsymbol{z}|\boldsymbol{l}\right) p_\theta \left(\boldsymbol{s}^f | \boldsymbol{z}\right)$, and the dashed lines represent $q_\phi \left(\boldsymbol{z} | \boldsymbol{s}^o\right)$. The intractable model $p_\theta \left(\boldsymbol{s}^f | \boldsymbol{z}\right)$ and $q_\phi \left(\boldsymbol{z} | \boldsymbol{s}^o\right)$ are represented by deep neural networks. The parameter $\psi$ for the priors is learned simultaneously with $\theta$ and $\phi$, which are the parameters of the variational inference.
	}
	\label{graph_model}
\end{figure*}
The goal of this paper is to approximate the intractable observation model for the object from its single view.
In order to exploit the single view that the mobile robot actually acquires from the observation, we let the algorithm to infer the variational likelihood of the full shape from a single view. We take advantage of variational auto-encoder (VAE) \cite{VAE}, and let the latent variables obtained from the observed object follow a tractable prior. In consequence, it is possible to perform the numerical analysis of the SLAM with data association problem using expectation maximization (EM) algorithm.
For the approximation of the observation distribution through the generative model, we introduce the Bayesian network similar to \cite{fei_bayesian}. The overview of the proposed method is shown in Fig.~\ref{likelihood_inference}.


Our contributions are two-fold: First, we show that the semantic features can be replaced by the variational latent variables for EM formulation of probabilistic SLAM; second, we introduce the encoding method for the generative model with the Bayesian networks of 3D object, exploiting its single view.

\section{Related Work}
Works on SLAM with features typically address the problem of data association.
In case of using an object as a feature, object recognition method based on learning with non-linear function is usually performed.
Therefore, obtaining a closed form solution is challenging since the posterior including data association follows intractable distributions.
To relax the problem, traditional approaches to SLAM divide the problem into front-end and back-end \cite{fastslam,graphslam}. In front-end, feature extraction and data association are solved relying on the detection algorithm such as object recognition. Subsequently, localization is performed using filter-based or graph optimization method in the back-end.
Due to this partitioned structure of SLAM problem, once the data association in the front-end fails, it is hard to avoid the tremendous error of the localization in the back-end. 

In order to overcome these issues, several methods have been developed to modify the false data association by weighting the association results \cite{back_end_graph,switchable}. However, proposed methods have limitations to deal with the uncorrelated data association and loop closing. \cite{PDASS} proposed a probabilistic data association formulation for semantic SLAM, using EM algorithm. They started with the pose optimization problem, and introduced the random variable of data association as latent one. Therefore, the approximated solution of maximum a posterior (MAP) for SLAM problem can be achieved despite the failure of initial data association.
However, the existing semantic SLAM basically performs data association based on object recognition algorithm such as \cite{faster_rcnn,yolo9000, active_deformable}, which usually has intractable observation probability distribution.
Thus even if the EM algorithm is used, the iterative solution of the object label is obtained in the expectation part by treating the label as a latent variable along with the data association.
Besides, this method hardly consider that observation of full shape with single mobile robot in real-time is infeasible.

Therefore, we show that the approximated observation model with variational inference enables the complete EM algorithm for the semantic SLAM problem.
In order to formulate the generative models for the variational inference, we assume that latent variables related to some factors such as class, instance categories and camera position are involved to create the object. In this aspect, our proposed model is similar to the studies of disentangled representation in \cite{generate_chairs,inverse_graph_conv}.
These works associated the latent variables with specific elements of the object or face such as contrast in image, object categories, facial angle, emotional expression, gender, and so on. 
In this way, those worked on how the objects are generated when latent variables corresponding to the specific elements change.

Since most of the objects are observed as a single view, we train the VAE to infer the variational likelihood of full shape from a single view of an object.
For actual implementation and evaluation, we assume that the single view obtained from a range finder or depth camera is represented as a voxel grid as well as full shape.

For the SLAM with data association problem, we will show that it is unnecessary to estimate the full shape of the object in actual practice as we only need the encoded features obtained from the encoder.
However, our network basically is an auto-encoder, thus it can be used as a shape retrieval network from a single view.
In this perspective, \cite{3D_GAN} is similar to our study.
\cite{vector_representation,marrnet,3D_rec_GAN} are also quite similar to our work in terms of the fact that they attempt to match a 3D shape to a single view, even though hardly consider the probabilistic approaches.

\section{Graphical Models for Likelihood of 3D Object}
Suppose the mobile robot observes the full shape of object, and represents it as voxel grid. In our work, we use this voxelized object shape as a semantic feature $\boldsymbol{s}^f$. 
To learn the generative model for estimating the observation model of 3D object, it is useful to introduce a Bayesian graphical model as \cite{fei_bayesian}. Similar to \cite{generate_chairs}, we assume that the Bayesian random process for the 3D object involves $\boldsymbol{l} = \{\boldsymbol{l}^c,\boldsymbol{l}^i,\boldsymbol{l}^v,\boldsymbol{l}^t\}$: $\boldsymbol{l}^c$-the category, $\boldsymbol{l}^i$-the characteristics of the observed instance, $\boldsymbol{l}^v$-camera viewpoint around the center of the object, and $\boldsymbol{l}^t$-translations in the voxel grid. The category $\boldsymbol{l}^c$ denotes the class of objects related to the rough appearance, and $\boldsymbol{l}^i$ stands for the detailed shape of the individual instance. We represent this Bayesian process as a directed acyclic graph model in Fig.~\ref{graph_model}(a).

For the generative model, the joint probability distribution can be denoted as $p\left(\boldsymbol{s}^f, \boldsymbol{l}\right) = p\left(\boldsymbol{l}\right)p\left(\boldsymbol{s}^f|\boldsymbol{l}\right)$. we simply assume that $p\left(\boldsymbol{l}\right) = p\left(\boldsymbol{l}^c\right)p\left(\boldsymbol{l}^i|\boldsymbol{l}^c\right)p\left(\boldsymbol{l}^v\right)p\left(\boldsymbol{l}^t\right)$ and each of the factorized terms follows a uniform distribution. Since the likelihood of the Bayesian process is too complex to handle, we resort to approximations using deep generative model instead. For our work, VAE with multi-layered neural networks is adopted \cite{VAE}. We display the graphical model with  variational latent variables in Fig.~\ref{graph_model}(b). Then the lower-bound $\mathcal{L}$ of the likelihood $p\left(\boldsymbol{s}^f|\boldsymbol{l}\right)$ is represented as follows:
\begin{align}
	\nonumber
	\mathcal{L}
	&
	\left(
		\theta, \phi, \psi ; \boldsymbol{s}^f, \boldsymbol{l}
	\right)
	\\
	\nonumber
	&=
	-KL
	\left(
		q_\phi
		\left(
			\boldsymbol{z} |\boldsymbol{s}^f
		\right)
		||
		p_\psi
		\left(
			\boldsymbol{z} |\boldsymbol{l}
		\right)
	\right)
	+
	\mathbb{E}_{\boldsymbol{z}^{l}}
	\left[
		\log p_\theta
		\left(
			\boldsymbol{s}^f | \boldsymbol{z}^{l}
		\right)
	\right]
	\\
	&\text{with } 
	\boldsymbol{z}^{l}
	\sim q_\phi
	\left(
		\boldsymbol{z} |\boldsymbol{s}^f
	\right).
	\label{lower_bound}
\end{align}
By the mean field inference \cite{VAE}\cite{ladderVAE}, we assume that $\boldsymbol{z}^{l}=
\left(
	\boldsymbol{z}^{c}, \boldsymbol{z}^{i},
	\boldsymbol{z}^{v}, \boldsymbol{z}^{t}
\right)$. Hence the variational likelihood can be factorized as
$q_\phi\left(\boldsymbol{z}|\boldsymbol{s}^f\right) = 
\prod_{\omega} 
q_{\phi^\omega}\left(\boldsymbol{z} |\boldsymbol{s}^f\right)$ for all symbol $\omega$ that $\boldsymbol{l}^\omega \in \boldsymbol{l}$. Similarly, we assume that the prior of $\boldsymbol{z}$ can also be factorized as the product of conditional densities such that
$
p_\psi\left(\boldsymbol{z}|\boldsymbol{l}\right)
=
\prod_{\omega}
p_\psi\left(\boldsymbol{z}|\boldsymbol{l}^\omega\right)
$. Then the KL-divergence term in \eqref{lower_bound} can be represented as follows:
\begin{align}
\nonumber
	KL
	&
	\left(
		q_\phi
		\left(
			\boldsymbol{z} |\boldsymbol{s}^f
		\right)
	||
		p_\psi
		\left(
			\boldsymbol{z} |\boldsymbol{l}
		\right)
	\right)
	\\
	\nonumber
	&=
	\sum_{\omega}
	KL
	\left(
		q_{\phi^\omega}
		\left(
			\boldsymbol{z} | \boldsymbol{s}^f
		\right)
		||
		p_\psi
		\left(
			\boldsymbol{z} | \boldsymbol{l}^\omega
		\right)
	\right).
\end{align}
In many studies of using VAE, the prior of $\boldsymbol{z}$ is simply assumed to be $p_\psi\left(\boldsymbol{z}\right) = \mathcal{N}\left(\boldsymbol{z}; \boldsymbol{0}, \boldsymbol{I}\right)$.
However, now the prior depends on $\boldsymbol{l}^{\omega}$, we let 
$p_{\psi}\left(\boldsymbol{z} | \boldsymbol{l}^\omega \right)
= \mathcal{N}\left(\boldsymbol{z}; 
\boldsymbol{\mu}^{{\omega}}, \boldsymbol{\Sigma}^{{\omega}}\right)
$ where $\left(\boldsymbol{\mu}^{{\omega}}, \boldsymbol{\Sigma}^{{\omega}}\right) = f_\psi\left(\boldsymbol{l}^\omega\right)$, i.e., nonlinear function with parameter $\psi$. Therefore we construct the neural networks for $f_\psi$ which denotes the prior distributions and also should be inferred from the training data. For simplicity, we can have 
$\boldsymbol{\Sigma}^{{\omega}} = \left(\boldsymbol{\sigma}^{{\omega}}\right)^2\boldsymbol{I}$,
which is a diagonal covariance matrix. For more simplicity we let $\boldsymbol{\Sigma}^{{\omega}}=\boldsymbol{I}$ and only leave $\boldsymbol{\mu}^\omega$ as trainable variables.
Similar to the prior,
we assume that $q_{\phi^\omega}\left(\boldsymbol{z}|\boldsymbol{s}^f\right)$ be the multivariate Gaussians with diagonal covariance 
$\mathcal{N}\left(\boldsymbol{z}; \boldsymbol{\mu}^{{\boldsymbol{s}\omega}}, \left(\boldsymbol{\sigma}^{{\boldsymbol{s}\omega}}\right)^2\boldsymbol{I} \right)$ as in \cite{VAE}.

Unlike our previous assumption, the mobile robot usually observes objects in the form of a single view such as an RGB-D image, thus obtaining the full shape is challenging for the real-time performance in practice. To infer the true observation model of the full shape, which is ideal for view-independent feature matching and loop closure, we exploit the single view and let variational likelihood indirectly infer the latent variables implying full shape from it. Therefore, similar to \cite{3D_GAN}, we redefine the  variational likelihood as $q_\phi\left(\boldsymbol{z}|\boldsymbol{s}^o\right)$, where $\boldsymbol{s}^o$ is an observed single view of an object. In this paper, $\boldsymbol{s}^o$ is assumed to be a voxelized grid which is converted from segmented depth images or point clouds.

\section{Variational Latent Variables and Semantic Features}
\subsection{Variational Latent Variables and Semantic SLAM with Data Association}
Consider the localization and mapping problem with object semantic features. As in \cite{PDASS}, assume that we have a collection $\mathcal{L} = \{ l_m=\left({l}_m^p, l_m^{c}\right) \}^M_{m=1}$ of $M$ static landmarks. The goal of the semantic SLAM is to estimate the 3D coordinate ${l}_m^p$ and label $l_m^{c}$ of the landmark, and robot poses $\mathcal{{X}} = \{\boldsymbol{x}_t\}^T_{t=1}$ given a set of object observations $\mathcal{{S}} = \{\mathcal{S}_t\}^T_{t=1}$. 
A $k$'th object detection 
$\boldsymbol{s}_k 
= 
(
	\boldsymbol{s}_k^p, \boldsymbol{s}_k^f
) \in \mathcal{S}_t$
from keyframe $t$ is composed of a full shape $\boldsymbol{s}^f$ and a 3D coordinate $\boldsymbol{s}^p$ for the center of its bounding box.
With the latent variables $\mathcal{{D}}$ for data association, the EM formulation for the semantic SLAM can be represented as follows:
\begin{align}
	w^t_{ij}
	&=
	\frac
	{
		\sum_{\mathcal{{D}}_t^\prime \in \mathbb{D}_t\left(i,j\right)}
		p\left(
			\mathcal{{S}}_t|\mathcal{{X}},\mathcal{L},\mathcal{D}_t
		\right)
	}
	{
		\sum_{\mathcal{D}_t \in \mathbb{D}_t}
		p\left(
		\mathcal{S}_t|\mathcal{X},\mathcal{L},\mathcal{D}_t
		\right)	
	}\text{ }\text{ }\text{ } \forall t,i,j
	\label{weight}
	\\
	\mathcal{X}, \mathcal{L}
	&=
	\argmin_{\mathcal{X}, \mathcal{L}}
	\sum_{t=1}^{T}
	\sum_{\boldsymbol{s}_k\in\mathcal{S}_t}
	\sum_{j}
	-w^t_{kj}
	\log p
	\left(
		\boldsymbol{s}_k | \boldsymbol{x}_t, l_j
	\right).
	\label{XL_argmin}
\end{align}
$\mathbb{D}_t$ is the set of all possible data associations 
$\mathcal{D}_t = \{\left(\alpha_k, \beta_k \right) \}^K_{k=1}$ representing that the object detection $\boldsymbol{s}_k$ of landmark $l_{\beta_k}$ was obtained from the robot state $\boldsymbol{x}_{\alpha_k}$.
Also,
$\mathbb{D}_t\left(i,j\right)\subseteq\mathbb{D}_t$ is the set of all possible data association $\mathcal{D}_t^\prime = \{\left(\alpha_k^\prime, \beta_k^\prime \right) \}$ such that $i$th detection is assigned to $j$th landmark.
For more details of the EM formulation, please refer to \cite{PDASS}.
Note that unlike \cite{PDASS}, we only set the data association $\mathcal{D}$ as latent variables except the object label $l^c$ for EM algorithm. 

Now assume that observation events are iid, i.e., $p\left(\mathcal{S}|\mathcal{X},\mathcal{L},\mathcal{D}\right)
= 
\prod_{k}
p(
	\boldsymbol{s}_k^p|\boldsymbol{x}_{\alpha_k}, l_{\beta_k}^p
)
p(
\boldsymbol{s}_k^{f}|l_{\beta_k}^{c}
)
$.
Here, if we set a landmark label $l^c = \left(\boldsymbol{l}^c, \boldsymbol{l}^i\right)$ and apply the lower bound in \eqref{lower_bound} to approximate the true likelihood $p\left(\boldsymbol{s}^f|l^c\right)$, 
the EM formulation in \eqref{weight} and \eqref{XL_argmin} can be represented as follows:
\begin{align}
	w^t_{ij}
	&\simeq
	\frac
	{
		\sum_{\mathcal{{D}}_t^\prime \in \mathbb{D}_t\left(i,j\right)}
		\prod_{k}
		p\left(
		\boldsymbol{s}^p_k|\boldsymbol{x}_{\alpha_k^\prime},l^p_{\beta_k^\prime}
		\right)		
		p_\psi\left(
		\boldsymbol{\mu}^{\boldsymbol{s}C}_k
		|
		l^c_{{\beta}^\prime_k}
		\right)
	}
	{
		\sum_{\mathcal{{D}}_t \in \mathbb{D}_t}
		\prod_{k}
		p\left(
		\boldsymbol{s}^p_k|\boldsymbol{x}_{\alpha_k},l^p_{{\beta}_k}
		\right)
		p_\psi\left(
		\boldsymbol{\mu}^{\boldsymbol{s}C}_k
		|
		l^c_{{\beta}_k}
		\right)
	}
	\label{weight_proposed}
	\\
	\mathcal{X}, \mathcal{L}
	&\simeq
	\argmin_{\mathcal{X}, \mathcal{L}}
	\sum_{t,k,j}
	-w^t_{kj}
	\log 
	p\left(\boldsymbol{s}^p_k|\boldsymbol{x}_t, l^p_j\right)
	p_\psi\left(\boldsymbol{\mu}^{\boldsymbol{s}C}_k | l^c_j \right),
	\label{XL_argmin_proposed}
\end{align}
where $\boldsymbol{\mu}^{\boldsymbol{s}C} 
= \left(\boldsymbol{\mu}^{\boldsymbol{s}c} ,\boldsymbol{\mu}^{\boldsymbol{s}i}  \right)$
and
$p_\psi\left(\boldsymbol{z}|l^c\right)
= 
p_\psi(\boldsymbol{z}|\boldsymbol{l}^c)
p_\psi(\boldsymbol{z}|\boldsymbol{l}^i)
$ (for more details, see Appendix I).
Consequently, when performing the EM with variational latent variables, we can simply replace $p\left({\boldsymbol{s}}^f|l^c\right)$ with 
$p_\psi(\boldsymbol{z}|l^c)$ which is a tractable Gaussian distribution, and substitute $\boldsymbol{\mu}^{\boldsymbol{s}C}$ which is the encoded feature from observed single view $\boldsymbol{s}^o$. 
In other words, the semantic feature $\boldsymbol{s}^f$ can be replaced with the encoded feature $\boldsymbol{\mu}^{\boldsymbol{s}C}$.
Therefore, even if the full shape $\boldsymbol{s}^f$ is hard to observe and its observation model is intractable, we can exploit the single view $\boldsymbol{s}^o$ to infer the true weight and optimal solutions for EM approximately.
\begin{figure*}[t]
	\centering
	\includegraphics[scale=0.15]{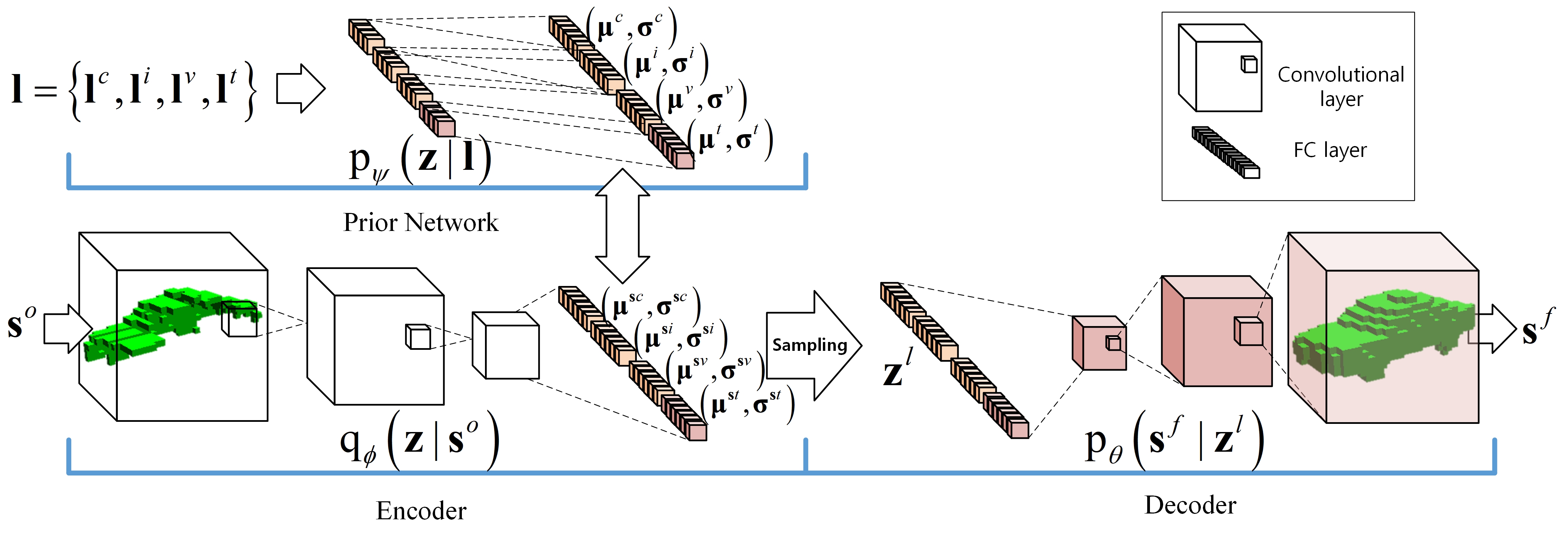}
	\caption{
		Proposed network architecture. The variational likelihood $q_\phi\left(\boldsymbol{z}|\boldsymbol{s}^o\right)$ and the inference distribution $p_\theta\left(\boldsymbol{s}^f|\boldsymbol{z}^l\right)$ are implemented by 3D convolutional layers and fully connected layers. The priors $p_\psi\left(\boldsymbol{z}|\boldsymbol{l}\right)$ of the variational latent variables $\boldsymbol{z}$ are also constructed with fully connected layers. The entire networks are trained end-to-end.
	}
	\label{proposed_networks}
\end{figure*}

\subsection{MLE with Variational Latent Variables}
In addition to the EM formulation, the variational lower bound can be used to solve the classification problem of the semantic features. Consider the maximum likelihood estimation (MLE) of $p\left(\boldsymbol{s}^f|l^c\right)$. With \eqref{lower_bound},  the approximated MLE can be denoted as follows:
\begin{align}
	\argmax_{
		l^c
	}
	p\left(
		\boldsymbol{s}^f|l^c
	\right)
	\simeq
	\argmax_{
		l^c
	}
	p_\psi
	\left( \boldsymbol{\mu}^{{\boldsymbol{s}}C}
		|
		l^c
	\right)
	\label{argmax_approximated}
\end{align}
(see Appendix II for the proof).
The approximated optimal solution is therefore obtained from the MLE with tractable variational prior $p_\psi\left(\boldsymbol{z}|l^c\right)$ and encoded feature $\boldsymbol{\mu}^{{\boldsymbol{s}}C}$.
As similar to the EM case, the classification solution of the full shape can be obtained from the MLE approximately since we utilize the single view to encode the features $\boldsymbol{\mu}^{\boldsymbol{s}C}$ which ultimately try to denote the full shape.
Note that the usage of 3D voxelized shape $\boldsymbol{s}$ as features is not a limitation of our method, and the algorithm can be applied to any other form of the features such as mesh grid or RGB-D images.

\section{Training Details}
\subsection{Data Augmentation}
In order to train and evaluate the proposed method, we use ModelNet10 and ModelNet40 datasets which are the subset of ModelNet 3D CAD datasets. 
For our experiments, two training sets are used; One is the single view data from 12 viewpoints of each object, and the other is the dataset from 24 viewpoints. 
Voxels of the single views are created through perspective projection.
During the training, random translations are performed for each sample as the same procedure done in \cite{voxnet}. In addition, two copies are used for all the samples and we add random noise to one of them.
Our graphical model involves not only the categories but also instance labels, translation and viewpoint, so that random flipping is not conducted since it can change the object's characteristics of instance and viewpoint index.


\subsection{Loss Function}
\subsubsection{Lower Bound}
In order to train the VAE, the negative variational lower bound  from \eqref{lower_bound} is used as the loss function. Since we assume that $q_\phi\left(\boldsymbol{z}|\boldsymbol{s}^o\right)$ and $p_\psi\left(\boldsymbol{z} |\boldsymbol{l}\right)$ be the multivariate Gaussians with diagonal covariance, the KL-divergence in \eqref{lower_bound} is expressed as follows:
\begin{align}
	\nonumber
	&KL
	\left(
		q_\phi
		\left(
			\boldsymbol{z} |\boldsymbol{s}^o
		\right)
		||
		p_\psi
		\left(
			\boldsymbol{z} |\boldsymbol{l}
		\right)
	\right)
	\\
	\nonumber
	&=
	\sum_{{\omega}}
	\sum_{j}
	\left(
		\log
		\frac{
			{\sigma}^{{\omega}}_j
		}
		{
			{\sigma}^{{\boldsymbol{s}}{\omega}}_j
		}
	+
	\frac{
		\left({\sigma}^{{\boldsymbol{s}}{\omega}}_j\right)^2
		+
		\left(
			{\mu}^{{\boldsymbol{s}}{\omega}}_j 
			- 
			{\mu}^{{\omega}}_j
		\right)^2
	}
	{
		2 \left({\sigma}_{j}^{\omega}\right)^2
	}
	-
	\frac{1}{2}
	\right),
\end{align}
where $\mu_j$ and $\sigma_j$ are the $j$th component of $\boldsymbol{\mu}$ and $\boldsymbol{\sigma}$, respectively.
The expectation term in \eqref{lower_bound} can be estimated by the reparameterization trick \cite{VAE} as follows:
\begin{align}
	\nonumber
	\mathbb{E}_{\boldsymbol{z}^{l}}
	\left[
		\log p_\theta
		\left(
			{\boldsymbol{s}}^f|\boldsymbol{z}^{l}
		\right)
	\right]
	\simeq
	\frac{1}{N}
	\sum_{n=1}^{N}
	\log p_\theta
	\left(
		{\boldsymbol{s}}^f|\boldsymbol{z}^{l}_n
	\right),
\end{align}
where $\boldsymbol{z}^{l} \sim q_\phi\left(\boldsymbol{z}|{\boldsymbol{s}}^o\right)$. Since the 3D shape of object is represented as binary random variables, we let $p_\theta \left({\boldsymbol{s}}^f|\boldsymbol{z}\right)$ be the Bernoulli distributions.

\subsubsection{regularization of the prior distribution}
When the class or instance labels of the objects are different, the encoded features $\boldsymbol{\mu}^{\boldsymbol{s}C}$ corresponding to those objects should also be different, which is ideal for solving EM or MLE. Therefore in chapter III, we introduce the prior $p_\psi\left(\boldsymbol{z}|\boldsymbol{l}\right)$ for each labels, and construct neural networks for nonlinear function $f_\psi$ which ouputs the mean $\boldsymbol{\mu}^\omega$
(note that the variance $\boldsymbol{\Sigma}^\omega$ is assumed to be $\boldsymbol{I}$).

\begin{figure*}[t]
	\centering
	\includegraphics[scale=0.25]{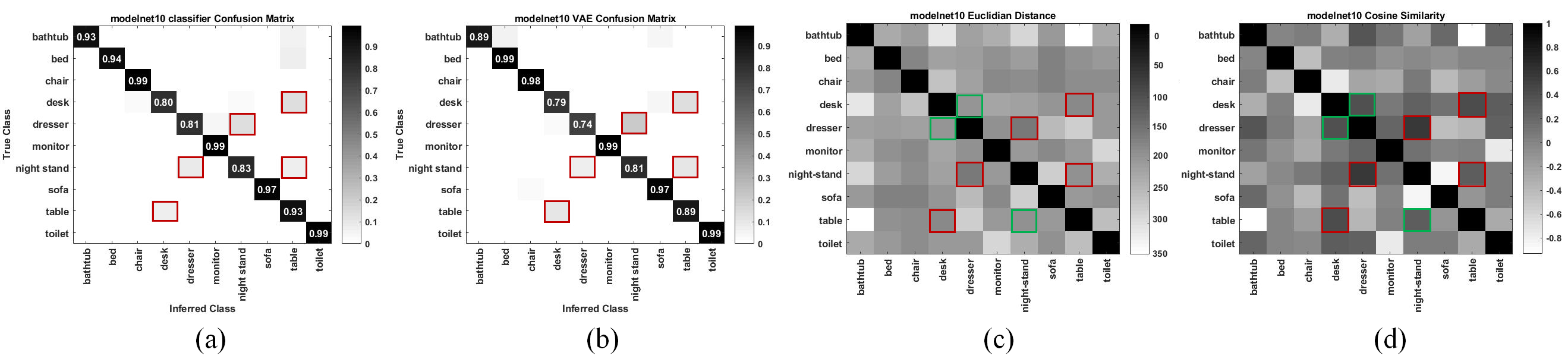}
	\caption{
		(a) The classification results of the classifier trained with full shapes and (b) the results of MLE with the proposed VAE model. (c) The Euclidian distance and (d) the cosine similarity between the parameter $\boldsymbol{\mu}^c$ of the prior networks are also displayed. We mark the common similar confusion pairs for (a)-(d) with red box. We also mark the pair of classes nearest to each other based on both Euclidian and cosine similarity in latent space with green box.		
		Even though our network has single views as inputs, it can find the relationships among the full shapes since the algorithm tries to infer the distributions of the full shape exploiting the object's single view. Therefore the results of the classificaions, MLE results and the distance matrices show the similar aspects. In addition, we can find similarities of the objects which are hardly observable in classification task. For example, in (c) and (d), we see that dresser and desk, and table and night stand class pairs have similar $\boldsymbol{\mu}^c$ as they have common properties on which we can put some stuffs.
	}
	\label{confusion matrix}
\end{figure*}
The problem is that the initial values of the weight and bias of the neural networks are close to $\boldsymbol{0}$ \cite{glorot}, so that the encoded features $\boldsymbol{\mu}^{\boldsymbol{s}C}$ for various objects have small variations at the beginning of learning. Moreover, $\boldsymbol{\mu}^\omega$ also has small variation regardless of the labels for a similar reason. Consequently even after the learning converges the performance of the EM and MLE is actually poor. Therefore, it is necessary to limit the distance between the prior means on the latent space to some threshold.
To do this, we define regularization loss $L^{rg}$ as follows:
\begin{align}
	\nonumber
	L^{rg}
	=
	\sum_{{\omega}}
	\sum_{\boldsymbol{l}^{\omega}_i,\boldsymbol{l}^{\omega}_j}
	l^{rg}\left(\boldsymbol{l}^{\omega}_i,\boldsymbol{l}^{\omega}_j \right) \text{ }\text{ }\forall i,j,
\end{align}
where

\begin{align}
	\nonumber
	l^{rg}&\left(\boldsymbol{l}^{\omega}_i,\boldsymbol{l}^{\omega}_j\right)
	\\
	\nonumber
	&=
	\begin{cases}
	\left(
		||
		{\boldsymbol{\mu}}^{\omega}_i
		-
		{\boldsymbol{\mu}}^{\omega}_j
		|| - \delta^\omega
	\right)^2
	&\text{if }
	||
	{\boldsymbol{\mu}}^{\omega}_i
	-
	{\boldsymbol{\mu}}^{\omega}_j
	|| < \delta^\omega
	\\	
	0
	&\text{else}
	\end{cases}.
\end{align}
The threshold
$\delta^\omega$ denotes the minimum distance in the latent space between means according to the different labels.

\subsubsection{reconstruction}
In the proposed algorithm, VAE can be regarded as a retrieval network from single view to full shape.
The expectation term and the KL-divergence term in \eqref{lower_bound} can also be regarded as the terms of shape retrieval and a regularization factor, respectively.
In our case, we find that when the KL-divergence term strongly restricts the network, it easily plunges into the local minima and gives the false results such as all-zero assigned voxels or centralized spherical bulb shape. Therefore, we change the range of the binary variables for target voxelized shape from $[0,1]$ to $[-1,2]$, to increase the scale of the gradient for shape retrieval loss as in \cite{generative}. The modified loss $L^{rc} = \sum_{j}l^{rc}_j$ for shape retrieval is then as follows:
\begin{align}
\nonumber
	l^{rc}_j = 
	\begin{cases}
	-2\log\left(s^p_{j}\right) + \log\left(1-s^p_{j}\right) 
	& \mbox{if }s^t_{j}=1 \\
	\log\left(s^p_{j}\right) - 2\log\left(1-s^p_{j}\right) 
	& \mbox{else if }s^t_{j}=0
	\end{cases}
\end{align}
where ${s}_j^{p}$ and ${s}_j^{t}$ are $j$th component of the binary occupancy variables for the predicted full shape $\boldsymbol{s}^{f,p}$ and the target(true) shape $\boldsymbol{s}^{f,t}$ of the object, respectively.
With this modification, our network can efficiently converge to the optimal state, which infers the restricted encoded features and correct full shapes from single views.

\subsection{Training Networks}
We construct our VAE with basic convolutional neural networks and dense layers.
The overview of the proposed network is displayed in Fig.~\ref{proposed_networks}.
The encoding part of the VAE is composed of 9 convolutional layers and 2 dense layers.
Similarly, 2 dense layers and 9 transposed convolutional layers are adopted to implement the decoder. For the prior distribution $p_\psi$, we construct 2 dense layers. We use the Batch normalization 
for efficient and stable learning. The dropout regularization 
is applied to the hidden layer to add noise to the latent variables. For nonlinearity, all layers use the exponential linear unit (ELU) 
except for the last layer, which adopts a sigmoid function for implementing the Bernoulli distribution. We use the Adam
for the optimizer of our VAE network.

\section{Experiments}
\subsection{Variational Latent Variable and Object Shape}
The latent variables $\boldsymbol{z}\sim q_\phi\left(\boldsymbol{z}|\boldsymbol{s}^o\right)= \mathcal{N}\left(\boldsymbol{z};\boldsymbol{\mu}^{\boldsymbol{s}}, \left(\boldsymbol{\sigma}^{\boldsymbol{s}}\right)^2\boldsymbol{I}\right)$ are sampled from the variational likelihood, which approximates the distribution $p\left(\boldsymbol{z}|\boldsymbol{s}^f, \boldsymbol{l}\right)$. 
Therefore the encoded features $\boldsymbol{\mu}^{\boldsymbol{s}}, \boldsymbol{\sigma}^{\boldsymbol{s}}$ have correlations with the full shape $\boldsymbol{s}^f$. For verification, we compare the MLE results of class label using \eqref{argmax_approximated} with the results of the classifier trained with full shape of objects. We use ModelNet10 dataset for these evaluations. 
The structure of the classifier is the same as the encoder of the VAE except for the last layer, which consists of an additional dense layer with softmax activation.
As shown in Fig.~\ref{confusion matrix}, the confusion matrices for MLE and full shape classification results show similar aspects.

Meanwhile, the prior networks trained together with auto-encoder represent the distributions of the latent variables, which imply the class, instance and observation position. Therefore the parameters of the prior also reflect the characteristics of each labels. To examine this we show the distance between the parameters, especially $\boldsymbol{\mu}^c$, in Fig.~\ref{confusion matrix}.
The prior of the class label is a distribution of the latent variable that determines the rough shape of the object. Therefore the closer the shapes of the classes are, the closer the distance in latent space between the prior parameter $\boldsymbol{\mu}^c$ is. As a result, the distances show similar aspects to the confusion matrices of full shapes.

\subsection{Classification and Reconstruction}
We also compare the MLE results with the classification results of state-of-the-art multi-view based classification algorithms in Table.~\ref{classification}. 
Our results show better performance than most of the other multi-view or voxelized full-shape based classification algorithms.
In addition, despite the simple architecture of the encoding layer, the proposed algorithm shows a competitive results to \cite{generative} which shows the best performance with single view by applying the deep ResNet structure (8 convolutional layers for the shallowest path, 45 convolutional layers for the deepest path and 2 dense layers).

\begin{table}[h]
	\caption{Comparision of the Classification Results}
	\label{classification}
	\begin{center}
		\begin{tabular}{c c c c c}
			\hline
			Methods & & ModelNet10 & & ModelNet40\\
			\hline
			DeepPano\cite{deepPano} & & 88.66\% & & 82.54\% \\
			Voxnet\cite{voxnet} & & 92.00\% & & 83.00\% \\
			3D-GAN\cite{3D_GAN} & & 91.00\% & & 83.30\% \\
			GIFT\cite{GIFT} & & 92.35\% & & 83.10\% \\
			PANORAMA-NN\cite{panorama} & & 91.12\% & & 90.70\% \\
			LightNet\cite{LightNet} & & 93.94\% & & 88.93\% \\
			VRN\cite{generative} (single view) & & - & & 88.98\% \\
			\hline
			proposed & & \multirow{2}*{91.35\%} & & \multirow{2}*{86.82\%}\\
			(single view) & & & & \\
			\hline
		\end{tabular}
	\end{center}
\end{table}

\begin{figure}[t]
	\centering
	\includegraphics[scale=0.20]{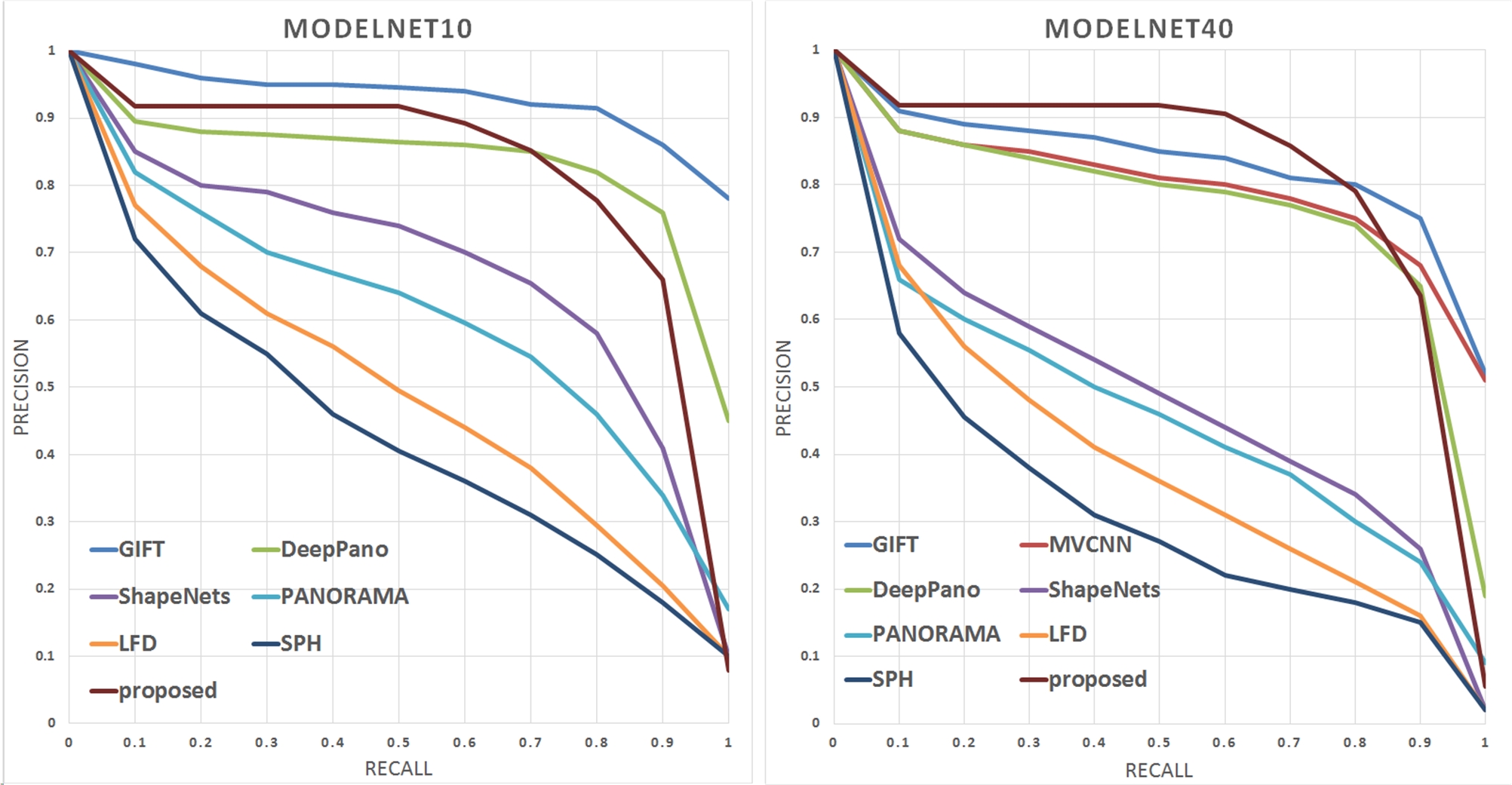}
	\caption{
		The precision-recall curve for (left) ModelNet10 and (right) ModelNet40 dataset. We include the results of LFD and SPH, which are the methods without neural networks. 
	}
	\label{precision_recall}
\end{figure}

\begin{figure*}[t]
	\centering
	\includegraphics[scale=0.175]{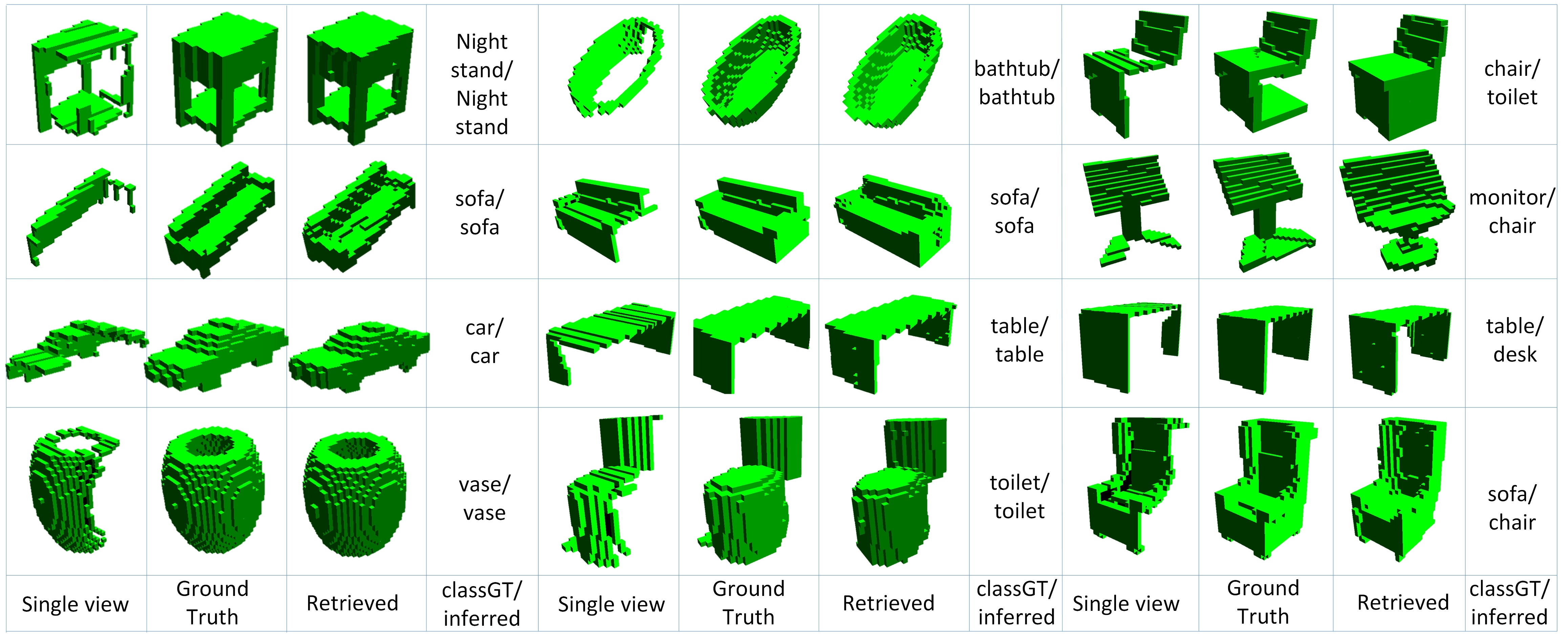}
	\caption{
		Examples of the shape retrieval and MLE results. We display the successful cases of classification on left and middle, and failure ones on right.
		In order to show the reconstructed parts, objects are arranged arbitrary regardless of the observation viewpoint from which the single view was obtained.
		Due to lack of the geometric information, shape retrieval fails in some cases such as in second row on the right side. Meanwhile, even if the shape inference is reasonably performed as shown in last row on the right side, MLE fails due to lack of the additional information such as material, color, size, etc. 
	}
	\label{retrieval_results}
\end{figure*}

The full shape of the observed single view can also be inferred from the trained autoencoder.
The retrieval results are shown in Table.~\ref{AUC_MAP},
and some samples of retrieval results are displayed in Fig.~\ref{retrieval_results}.
We also report the precision-recall curve in Fig.~\ref{precision_recall}.
Interestingly, unlike the classification results, the accuracy of retrieval results for the ModelNet10 and ModelNet40 dataset are not significantly different.
This is because some classes such as airplanes or cars that have distinctive shape features are easier to reconstruct than other furniture classes.

Meanwhile, since the scale, color, or material of the object is not considered at all during learning, object reasoning has limitations only by the geometric information from the single views.
For example, in the last line on the right side of Fig.~\ref{retrieval_results}, the full shape is reasonably inferred from the single view.
However, due to lack of the other information, the MLE result implies that it is the most likely to be the chair, not the sofa.
\begin{table}[h]
	\caption{Comparison of the Shape Retrieval Results}
	\label{AUC_MAP}
	\begin{center}
		\begin{tabular}{c c c c c c}
			\hline
			Methods & \multicolumn{2}{c}{ModelNet10} & &\multicolumn{2}{c}{ModelNet40}\\
			\cline{2-3}\cline{5-6}
			& AUC & mAP & & AUC & mAP\\
			\hline
			PANORAMA-NN\cite{panorama} & - & 87.39\% & & - & 83.45\% \\
			ShapeNets\cite{ShapeNet} & 69.28\% & 68.26\% & & 49.94\% & 49.23\% \\
			DeepPano\cite{deepPano} & 85.45\% & 84.18\% & & 77.63\% & 76.81\% \\
			GIFT\cite{GIFT} & 92.35\% & 91.12\% & & 83.10\% & 81.94\% \\
			\hline
			proposed & \multirow{2}*{81.94\%} & \multirow{2}*{82.72\%} & & \multirow{2}*{84.94\%} & \multirow{2}*{83.82\%}\\
			(single view) & & && &\\
			\hline
		\end{tabular}
	\end{center}
\end{table}

\section{CONCLUSION}
Since the high-dimensional feature such as the 3D shape of the object follows intractable observation model, numerical analysis for Bayesian inference such as semantic SLAM becomes challenging. To overcome this problem, we show that the semantic features can be replaced by variational latent variables. We also present a feature encoding method using the variational generative model. 
Since observing the full shape in real-time is challenging, the proposed algorithm infers the variational likelihood of the full shape from the single view. Therefore, complex observation model of 3D object is approximated to the tractable distribution.
Consequently, the encoded features and their priors enable the numerical analysis for the probabilistic estimation. 
Experiments are conducted to evaluate the algorithm on the 3D CAD dataset.
To analyze the approximated distributions and encoded features, we perform classification with maximum likelihood estimation, and shape retrieval by decoding process.


\section*{Appendix I : Variational Lower Bound and EM Algorithm}
As we assume that $p\left(\boldsymbol{l}\right)= p\left(\boldsymbol{l}^c\right)p\left(\boldsymbol{l}^i|\boldsymbol{l}^c\right)p\left(\boldsymbol{l}^v\right)p\left(\boldsymbol{l}^t\right)$ and each of the factorized priors is a uniform distribution,
the term $p(\boldsymbol{s}^f|l^c)$ for the observation model can be represented as
$p(\boldsymbol{s}^f|l^c) 
= 
a
\sum_{\boldsymbol{l}^v,\boldsymbol{l}^t}
p\left(
\boldsymbol{s}^f|\boldsymbol{l}
\right)$ where $a =
\frac{p\left(\boldsymbol{l}\right)}{p\left(l^c\right)}  = p\left(\boldsymbol{l}^v\right)p\left(\boldsymbol{l}^t\right)$ is constant.
Therefore, we can apply the variational lower bound to approximate the likelihood  $p\left(\boldsymbol{s}^f|\boldsymbol{l}\right)$ for observation model in EM formulation.
Since we are focusing on the likelihood of $l^c$, in the remaining part we let 
$p_\psi\left(\boldsymbol{z}|l^c\right) 
= 
p_\psi\left(\boldsymbol{z}|\boldsymbol{l}^c\right)
p_\psi\left(\boldsymbol{z}|\boldsymbol{l}^i\right)$
and
$q_{\phi^C}\left(\boldsymbol{z}|\boldsymbol{s}^o\right)
=
q_{\phi^c}\left(\boldsymbol{z}|\boldsymbol{s}^o\right)
q_{\phi^i}\left(\boldsymbol{z}|\boldsymbol{s}^o\right)
$ for convenience.

\subsection{Variational Lower Bound for Expectation Step}
After the convergence of VAE training, the variational lower bound in \eqref{lower_bound} approximates the loglikelihood of $\boldsymbol{s}^f$ $\log p\left(\boldsymbol{s}^f | \boldsymbol{l}\right)$ in \eqref{weight}. Therefore, we can rewrite the term $p\left(\boldsymbol{s}^f|l^c\right)$ as the following:
\begin{align}
	\nonumber
	p\left( \boldsymbol{s}^f|l^c \right)
	&\simeq
	a
	\sum_{\boldsymbol{l}^v,\boldsymbol{l}^t}
	\exp
	\left(
		\mathcal{L}
			\left(
				\theta, \phi, \psi ; \boldsymbol{s}^f, \boldsymbol{l}
			\right)
	\right)
	\\
	\nonumber
	&=
	a
	\sum_{\boldsymbol{l}^v,\boldsymbol{l}^t}
	\exp
	\left(
		\mathbb{E}_{\boldsymbol{z}^{l}}
		\left[
			\log p_\theta
			\left(
				\boldsymbol{s}^f | \boldsymbol{z}^{l}
			\right)
		\right]
	\right)
	\cdot
	\\
	\nonumber
	&\prod_{\omega}
	\exp
	\left(
		-KL
		\left(
			q_{\phi^{\omega}}
			\left(
				\boldsymbol{z} | \boldsymbol{s}^o
			\right)
			||
			p_\psi
			\left(
				\boldsymbol{z} | \boldsymbol{l}^\omega
			\right)
		\right)
	\right)
	\\
	&=
	a
	\kappa_e\left(\boldsymbol{s}^f\right)
	\kappa_{kl}\left(\boldsymbol{s}^o\right)
	\kappa_{kl}\left(\boldsymbol{s}^o, l^c\right),
	\label{approximated_likelihood}
\end{align}
where
\begin{align}
	\nonumber
	\kappa_e\left(\boldsymbol{s}^f\right)
	&=
	\exp
	\left(
		\mathbb{E}_{\boldsymbol{z}^{l}}
		\left[
			\log p_\theta
			\left(
				\boldsymbol{s}^f | \boldsymbol{z}^{l}
			\right)
		\right]
	\right)
	\\
	\nonumber
	\kappa_{kl}\left(\boldsymbol{s}^o\right)
	&=
	\sum_{\boldsymbol{l}^v}
	\exp
	\left(
		-KL
		\left(
			q_{\phi^{v}}
			\left(
				\boldsymbol{z} | \boldsymbol{s}^o
			\right)
			||
			p_\psi
			\left(
				\boldsymbol{z} | {\boldsymbol{l}^v}
			\right)
		\right)
	\right)
	\cdot
	\\
	\nonumber
	&\sum_{\boldsymbol{l}^t}
	\exp
	\left(
		-KL
		\left(
			q_{\phi^{t}}
			\left(
				\boldsymbol{z} | \boldsymbol{s}^o
			\right)
			||
			p_\psi
			\left(
				\boldsymbol{z} | {\boldsymbol{l}^t}
			\right)
		\right)
	\right)
	\\
	\nonumber
	\kappa_{kl}\left(\boldsymbol{s}^o, l^c\right)
	&=
	\exp
	\left(
		-KL
		\left(
			q_{\phi^{C}}
			\left(
				\boldsymbol{z} | \boldsymbol{s}^o
			\right)
			||
			p_\psi
			\left(
				\boldsymbol{z} | {l^c}
			\right)
		\right)
	\right).
\end{align}
Substituting \eqref{approximated_likelihood} into \eqref{weight} yields:
\begin{align}
\nonumber
&w^t_{ij}
\\
\nonumber
&\simeq
\frac
{
	\sum
	\prod_{k}
	p\left(
		\boldsymbol{s}^p_k|\boldsymbol{x}_{\alpha_k^\prime}, l^p_{\beta_k^\prime}
	\right)
	a
	\kappa_e\left(\boldsymbol{s}^f_k\right)
	\kappa_{kl}\left(\boldsymbol{s}^o_k \right)
	\kappa_{kl}\left(\boldsymbol{s}^o_k, l^c_{\beta_k^\prime}\right)
}
{
	\sum
	\prod_{k}
	p\left(
		\boldsymbol{s}^p_k|\boldsymbol{x}_{\alpha_k},l^p_{{\beta}_k}
	\right)
	a
	\kappa_e\left(\boldsymbol{s}^f_k\right)
	\kappa_{kl}\left(\boldsymbol{s}^o_k \right)
	\kappa_{kl}\left(\boldsymbol{s}^o_k, l^c_{{\beta}_k}\right)
}.
\end{align}
Since $a$, $\kappa_e\left(\boldsymbol{s}^f_k\right)$ and $\kappa_{kl}\left(\boldsymbol{s}^o_k \right)$ are independent to 
$\mathcal{D}_t = \{\left(\alpha_k,\beta_k\right) \} \subseteq \mathbb{D}_t$ and $\mathcal{D}_t^\prime=\{\left(\alpha_k^\prime,\beta_k^\prime\right) \} \subseteq \mathbb{D}_t\left(i,j\right)$, we can reduce the fraction as:
\begin{align}
w^t_{ij}
\simeq
\frac
{
	\sum_{\mathcal{D}_t^\prime \in \mathbb{D}_t\left(i,j\right)}
	\prod_{k}
	p\left(
	\boldsymbol{s}^p_k|\boldsymbol{x}_{\alpha_k^\prime},l^p_{\beta_k^\prime}
	\right)		
	\kappa_{kl}\left(\boldsymbol{s}^o_k, l^c_{\beta_k^\prime}\right)
}
{
	\sum_{\mathcal{{D}}_t \in \mathbb{D}_t}
	\prod_{k}
	p\left(
	\boldsymbol{s}^p_k|\boldsymbol{x}_{\alpha_k},l^p_{{\beta}_k}
	\right)
	\kappa_{kl}\left(\boldsymbol{s}^o_k, l^c_{{\beta}_k}\right)
}.
\label{weight_expand}
\end{align}
Focusing on the term $\kappa_{kl}\left(\boldsymbol{s}^o, l^c\right)$, we now expand the negative KL-divergence term as:
\begin{align}
	\nonumber
	-KL
	&\left(
		q_{\phi^{C}}
		\left(
			\boldsymbol{z} | \boldsymbol{s}^o
		\right)
		||
		p_\psi
		\left(
			\boldsymbol{z} | {l^c}
		\right)
	\right)	
	\\
	\nonumber
	&=
	\mathbb{E}_{\boldsymbol{z}}
	\left[
		\log p_\psi\left(
			\boldsymbol{z}|l^c
		\right)
	\right]	
	+
	H\left(
		q_{\phi^{C}}\left(
			\boldsymbol{z}|\boldsymbol{s}^o
		\right)
	\right).
\end{align}
Note that since we assume the priors of the variational latent variables are multivariate Gaussians with diagonal covariances in Section III, we can represent $p_\psi\left(\boldsymbol{z}|l^c\right)$
as
$
\mathcal{N}\left(
\boldsymbol{z}; 
\boldsymbol{\mu}^{C}, \boldsymbol{\Sigma}^{C}
\right)
$,
where
$\boldsymbol{\mu}^{C} 
= 
\left(\boldsymbol{\mu}^{c}, \boldsymbol{\mu}^{i}\right)$
,
$
\boldsymbol{\Sigma}^{C} = \left(\boldsymbol{\sigma}^C\right)^2\boldsymbol{I}
$
and
$\boldsymbol{\sigma}^{C} 
= 
\left(
\boldsymbol{\sigma}^{c}, \boldsymbol{\sigma}^{i}
\right)$.
Similarly,
we let
$q_{\phi^{C}}\left(\boldsymbol{z}|\boldsymbol{s}^o \right)
=
\mathcal{N}\left(
\boldsymbol{z}; 
\boldsymbol{\mu}^{\boldsymbol{s}C}, \boldsymbol{\Sigma}^{\boldsymbol{s}C}
\right)
$,
where
$\boldsymbol{\mu}^{\boldsymbol{s}C} 
= 
\left(\boldsymbol{\mu}^{\boldsymbol{s}c}, \boldsymbol{\mu}^{\boldsymbol{s}i}\right)$
,
$\boldsymbol{\Sigma}^{\boldsymbol{s}C} = \left(\boldsymbol{\sigma}^{\boldsymbol{s}C}\right)^2\boldsymbol{I}$
and
$\boldsymbol{\sigma}^{\boldsymbol{s}C} 
= 
\left(
\boldsymbol{\sigma}^{\boldsymbol{s}c}, \boldsymbol{\sigma}^{\boldsymbol{s}i}
\right)$.
Then we continue:
\begin{align}
	\nonumber
	&-KL\left(
		q_{\phi^{C}}
		\left(
		\boldsymbol{z} | \boldsymbol{s}^o
		\right)
		||
		p_\psi
		\left(
		\boldsymbol{z} | {l^c}
		\right)
	\right)
	\\
	\nonumber
	&=	
	\mathbb{E}_{\boldsymbol{z}}
	\left[
		-\log Z
		-\frac{1}{2}||\boldsymbol{z} - \boldsymbol{\mu}^{C}||^2_{\left(
			{\boldsymbol{\Sigma}}^{C}
			\right)^{-1}}
	\right]	
	+
	H\left(
		q_{\phi^{C}}\left(
		\boldsymbol{z}|\boldsymbol{s}^o
		\right)
	\right)
	\\
	\nonumber
	&=
	-\log Z
	-\frac{1}{2}
	||
		\boldsymbol{\mu}^{\boldsymbol{s}C} - \boldsymbol{\mu}^{C}
	||^2_{\left({\boldsymbol{\Sigma}}^{C}\right)^{-1}}
	-
	\frac{1}{2}
	\sum_{n}
	\left(
		\frac{\sigma^{\boldsymbol{s}C}_n}
		{\sigma^{C}_n}
	\right)^2
	+
	H,
\end{align}
where $Z$ is the normalization term for $p_\psi\left(\boldsymbol{z}|l^c\right)$.
Since there is no constraint to ${\boldsymbol{\sigma}}^{C}$, we simply let ${\sigma}^{C}_n = 1$ (see Section III).
The term $\kappa_{kl}\left(\boldsymbol{s}^o, l^c\right)$ then can be rewritten as:
\begin{align}
	\nonumber
	&\kappa_{kl}\left(\boldsymbol{s}^o, l^c\right)
	\\
	\nonumber
	&=
	\exp
	\left(
		-KL
		\left(
			q_{\phi^{C}}
			\left(
				\boldsymbol{z} | \boldsymbol{s}^o
			\right)
			||
			p_\psi
			\left(
				\boldsymbol{z} | {l^c}
			\right)
		\right)
	\right)
	\\
	\nonumber
	&=
	\frac{1}{Z}
	\exp\left(
		-\frac{1}{2}
		||
		\boldsymbol{\mu}^{\boldsymbol{s}C} - \boldsymbol{\mu}^{C}
		||^2_{\left({\boldsymbol{\Sigma}}^{C}\right)^{-1}}
	\right)
	\frac
	{\exp\left(
			H\left(
				q_{\phi^C}\left(
					\boldsymbol{z}|\boldsymbol{s}^o
				\right)
			\right)
		\right)}
	{\exp\left(
		\frac{1}{2}
		\sum_{n}
		\left(
			\frac{\sigma^{\boldsymbol{s}C}_n}
			{\sigma^{C}_n}
		\right)^2
	\right)
	}
	\\
	&=
	p_\psi\left(
		\boldsymbol{\mu}^{\boldsymbol{s}C}
		|
		l^c
	\right)
	\frac
	{\exp\left(
			H\left(
				q_{\phi^{C}}\left(
					\boldsymbol{z}|\boldsymbol{s}^o
				\right)
			\right)
		\right)}
	{\exp\left(
		\frac{1}{2}
		\sum_{n}
			\left(
				{\sigma^{\boldsymbol{s}C}_n}
			\right)^2
		\right)
	}.
	\label{KL_expand_exp}
\end{align}
As the exponential terms in \eqref{KL_expand_exp} is dependent not on the data association $\mathcal{D}$ but on $\boldsymbol{s}^o$, substituting \eqref{KL_expand_exp} into \eqref{weight_expand} and reducing the fraction finally yield:
\begin{align}
\nonumber
w^t_{kj}
\simeq
\frac
{
	\sum_{\mathcal{D}_t^\prime \in \mathbb{D}_t\left(k,j\right)}
	\prod_{k}
	p\left(
	\boldsymbol{s}^p_k|\boldsymbol{x}_{{\alpha}^\prime_k},l^p_{{\beta}_k}
	\right)		
	p_\psi\left(
	\boldsymbol{\mu}^{\boldsymbol{s}C}_k
	|
	l^c_{{\beta}^\prime_k}
	\right)
}
{
	\sum_{\mathcal{{D}}_t \in \mathbb{D}_t}
	\prod_{k}
	p\left(
	\boldsymbol{s}^p_k|\boldsymbol{x}_{{\alpha}_k},l^p_{{\beta}_k}
	\right)
	p_\psi\left(
	\boldsymbol{\mu}^{\boldsymbol{s}C}_k
	|
	l^c_{{\beta}_k}
	\right)
}.
\end{align}

\subsection{Variational Lower Bound for Maximization Step}
Similar to the expectation step, we can also apply the variational likelihood for the maximization step. Since we assumed that $p\left(\boldsymbol{s}|\boldsymbol{x},l\right) 
= 
p\left(\boldsymbol{s}^p|\boldsymbol{x}, l^p\right)
p\left(\boldsymbol{s}^f|l^c\right)$, \eqref{XL_argmin} can be rewritten as follows:
\begin{align}
\nonumber
\mathcal{X}, \mathcal{L}
=
\argmin_{\mathcal{X}, \mathcal{L}}
\sum_{t,k,j}
-w^t_{kj}
\log 
p\left(\boldsymbol{s}^p_k|\boldsymbol{x}_t, l^p_j\right)
p\left(\boldsymbol{s}^f_k|l^c_j\right).
\end{align}
Note that only $\kappa_{kl}\left(\boldsymbol{s}^o, l^c\right)$ and $p_\psi\left(\boldsymbol{\mu}^{\boldsymbol{s}C} | l^c \right)$ in \eqref{approximated_likelihood} and \eqref{KL_expand_exp} are related to $\mathcal{L} = \{\left({l}_m^p, l_m^{c}\right) \}$.
Therefore substituting \eqref{approximated_likelihood} and \eqref{KL_expand_exp} subsequently, we have:
\begin{align}
	\nonumber
	\mathcal{X}, \mathcal{L}
	=
	\argmin_{\mathcal{X}, \mathcal{L}}
	\sum_{t,k,j}
	-w^t_{kj}
	\log 
	p\left(\boldsymbol{s}^p_k|\boldsymbol{x}_t, l^p_j\right)
	p_\psi\left(\boldsymbol{\mu}^{\boldsymbol{s}C}_k | l^c_j \right)
	.
\end{align}
Since $\mathcal{L} = \{\left( l^p_m, l^c_m \right)  \}^M_{m=1}$, we can further expand the equation as follows:
\begin{align}
	\nonumber
	\mathcal{X}, l^p
	&=
	\argmin_{\mathcal{X}, l^p}
	\sum_{t,k,j}
	-w^t_{kj}
	\log 
	p\left(\boldsymbol{s}^p_k|\boldsymbol{x}_t, l^p_j\right),
	\\
	\nonumber
	l^c
	&=
	\argmin_{l^c}
	\sum_{t,k,j}
	-w^t_{kj}
	\log
	p_\psi\left(\boldsymbol{\mu}^{\boldsymbol{s}C}_k | l^c_j \right).
\end{align}

\section*{Appendix II : MLE with Variational Latent Variables}
Consider the MLE problem of $p\left({\boldsymbol{s}}^f|l^c\right)$. With the proposed method, the true likelihood of ${\boldsymbol{s}}^f$ is approximated to variational lower bound as \eqref{approximated_likelihood}. Hence the approximated solution for MLE  can be denoted as follows:
\begin{align}
\nonumber
\argmax_{
	l^c
}
p\left(
{\boldsymbol{s}}^f|l^c
\right)
&\simeq
\argmax_{
	l^c
}
a
\sum_{\boldsymbol{l}^v,\boldsymbol{l}^t}
\exp
\left(
\mathcal{L}
\left(
\theta, \phi, \psi ; \boldsymbol{s}^f, \boldsymbol{l}
\right)
\right)
\\
\nonumber
&=
\argmax_{
	l^c
}
a
\kappa_e\left(\boldsymbol{s}^f\right)
\kappa\left(\boldsymbol{s}^o \right)
\kappa\left(\boldsymbol{s}^o, l^c\right)
\\
\nonumber
&=
\argmax_{
	l^c
}
\kappa\left(\boldsymbol{s}^o, l^c\right).
\end{align}
Substituting \eqref{KL_expand_exp} finally yields:
\begin{align}
	\nonumber
	\argmax_{
		l^c
	}
	p\left(
		{\boldsymbol{s}}^f|l^c
	\right)
	&\simeq
	\argmax_{
		l^c
	}
	p_\psi\left(
	\boldsymbol{\mu}^{\boldsymbol{s}C}
	|
	l^c
	\right)
	\frac
	{\exp\left(
		H\left(
		q_{\phi^{C}}
		\right)
		\right)}
	{\exp\left(
		\frac{1}{2}
		\sum
			\left(
				{\sigma^{\boldsymbol{s}C}_n}
			\right)^2
		\right)}
	\\
	\nonumber
	&=
	\argmax_{
		l^c
	}
	p_\psi\left(
	\boldsymbol{\mu}^{\boldsymbol{s}C}
	|
	l^c
	\right).
\end{align}





\bibliographystyle{IEEEtran}
\bibliography{root_2018}

\end{document}